\documentclass[]{TEAI}
\usepackage{helvet}

\usepackage{amsmath} 
\usepackage{natbib}
\usepackage{graphicx}
\usepackage{subcaption} 

\usepackage[toc,page,header]{appendix}
\usepackage[utf8]{inputenc} 
\usepackage[T1]{fontenc}    

\usepackage{url}            
\usepackage{booktabs}       
\usepackage{lmodern}        
\usepackage{amsfonts}       
\usepackage{nicefrac}       
\usepackage{microtype}      
\usepackage{wrapfig}

\usepackage{amssymb}  
\usepackage{fontawesome}  
\usepackage{url}  

\usepackage{titletoc}

\usepackage{tikz}  
\usepackage{comment}  
\usepackage{tabularx}  
\usepackage{booktabs}  

\usepackage{minitoc}

\usepackage{booktabs}
\usepackage{array}
\usepackage{etoolbox}

\definecolor{lightblue}{RGB}{200, 230, 255}  
\definecolor{headerblue}{RGB}{150, 200, 255} 

\usepackage{pgfplots}
\usepackage[utf8]{inputenc} 
\usepackage[T1]{fontenc}    

\usepackage{url}            
\usepackage{booktabs}       
\usepackage{amsfonts}       
\usepackage{nicefrac}       
\usepackage{microtype}      
\usepackage{xcolor}         
\usepackage{graphicx}
\usepackage{float}
\usepackage{comment}
\usepackage{multirow} 
\usepackage{amsmath} 
\usepackage{makecell} 
\usepackage{siunitx}  
\usepackage{tikz}
\usepackage{pgf-pie} 
\usepackage{subcaption}
\usepackage{wrapfig}
\usepackage[export]{adjustbox}

\usepackage{ragged2e}      
\usepackage{tabularx}       
\usepackage{array}          
\usepackage{caption}        
\usepackage{enumitem}
\usepackage{pifont}
\usepackage[hang,flushmargin]{footmisc} 

\usepackage{tcolorbox}

\usepackage{tcolorbox}    
\tcbuselibrary{breakable}  
\tcbuselibrary{skins}      

\usepackage{tabularx}
\usepackage{listings}

\usepackage{times}
\usepackage{latexsym}

\usepackage[T1]{fontenc}

\usepackage[utf8]{inputenc}

\usepackage{microtype}

\usepackage{inconsolata}

\usepackage{graphicx}

\usepackage{amsmath}
\usepackage{amsfonts}

\usepackage[most]{tcolorbox}
\usepackage{listings}
\usepackage{ragged2e}
\usepackage{enumitem}
\usepackage{booktabs}
\usepackage[most]{tcolorbox}
\usepackage{listings}
\usepackage{ragged2e}
\tcbuselibrary{breakable,listings}

\usepackage{algorithm}
\usepackage{algorithmic}
\setboolean{ALC@noend}{true}
\usepackage{cuted} 

\definecolor{codegreen}{rgb}{0,0.6,0}
\definecolor{codegray}{rgb}{0.5,0.5,0.5}
\definecolor{codepink}{RGB}{252, 142, 172}
\definecolor{codepurple}{rgb}{0.58,0,0.82}
\definecolor{backcolour}{RGB}{245,245,245}
\lstdefinestyle{scaler}{
    backgroundcolor=\color{backcolour},   
    commentstyle=\color{magenta},
    keywordstyle=\color{blue},
    numberstyle=\tiny\color{codegray},
    stringstyle=\color{codepurple},
    basicstyle=\fontfamily{\ttdefault}\footnotesize,
    breakatwhitespace=false,        
    breaklines=true,                
    keepspaces=true,    
    frame=single,
    numbersep=5pt,                  
    showspaces=false,              
    showstringspaces=false,
    showtabs=false,               
    tabsize=2,
    classoffset=1, 
    keywordstyle=\color{violet},
    classoffset=0,
}
\newtcblisting{PromptBox}[1]{%
enhanced,
  breakable,
  listing only,
  colback=white,
  colframe=black!60,
  boxrule=0.8pt,
  arc=2pt,
  left=6pt,right=6pt,top=6pt,bottom=6pt,
  colbacktitle=black!60,
  coltitle=white,
  fonttitle=\bfseries,
  title={#1},
  before upper=\RaggedRight,
  segmentation style={draw=none},
  listing engine=listings,
  listing options={style=prompt}
}
\usepackage{amsmath}
\usepackage{array}
\usepackage{longtable} 

\title{\textsc{SCALER}: Synthetic Scalable Adaptive Learning Environment for Reasoning}

\author{
 \textbf{Caijun Xu\textsuperscript{1,2}},
 \textbf{Changyi Xiao\textsuperscript{1}},
 \textbf{Zhongyuan Peng\textsuperscript{1}},
 \textbf{Xinrun Wang\textsuperscript{3}},
 \textbf{Yixin Cao\textsuperscript{1,2,$\dagger$}},
}
\affiliation[1]{\mbox{Fudan University}} 
\affiliation[2]{\mbox{Shanghai Innovation Institute}}
\affiliation[3]{\mbox{Singapore Management University}}

\abstract{
\begin{abstract}

Reinforcement learning (RL) offers a principled way to enhance the reasoning capabilities of large language models, yet its effectiveness hinges on training signals that remain informative as models evolve. 
In practice, RL progress often slows when task difficulty becomes poorly aligned with model capability or when training is dominated by a narrow set of recurring problem patterns.
To jointly address these issues, we propose \textbf{SCALER} (\textbf{S}ynthetic s\textbf{C}alable \textbf{A}daptive \textbf{L}earning \textbf{E}nvironment for \textbf{R}easoning), a framework that sustains effective learning signals through adaptive environment design.
SCALER introduces a scalable synthesis pipeline that converts real-world programming problems into verifiable reasoning environments with controllable difficulty and unbounded instance generation, enabling RL training beyond finite datasets while preserving strong correctness guarantees. Building on this, SCALER further employs an adaptive multi-environment RL strategy that dynamically adjusts instance difficulty and curates the active set of environments to track the model’s capability frontier and maintain distributional diversity. This co-adaptation prevents reward sparsity, mitigates overfitting to narrow task patterns, and supports sustained improvement throughout training. Extensive experiments show that SCALER consistently outperforms other RL baselines across diverse reasoning benchmarks and exhibits more stable, long-horizon training dynamics.
\end{abstract}
}

\correspondence{\email{cjxu25@m.fudan.edu.cn}}
\checkdata[Github]{\url{https://github.com/ALEX-nlp/SCALER}}

\begin{document}
\maketitle
\renewcommand{\thefootnote}{}
\footnotetext{$^*$Equal Contribution.\\$^\dagger$Corresponding authors.}
\renewcommand{\thefootnote}{\arabic{footnote}}


\vspace{-1.5em}

\section{Introduction}
\label{sec:intro}

Reinforcement learning (RL) has become a key post-training paradigm for enhancing the reasoning capabilities of large language models (LLMs) ~\cite{openai2024openaio1card,deepseekai2025deepseekr1incentivizingreasoningcapability}. By optimizing a policy under explicit, verifiable rewards, RL can sharpen decision making, improve long-horizon credit assignment, and expand the model's reasoning frontier~\cite{shao2024deepseekmath,yu2025dapoopensourcellmreinforcement,hu2025reinforce++}. 
However, scaling RL for reasoning is often bottlenecked not by the optimizer itself, but by the availability of continuously effective reward signals throughout training~\cite{razin2023vanishing,razin2025makes,zhang2025confclipconfidenceweightedclippedreward}.

In this work, we argue that the training signal, for RL to continue improving LLMs, should remain effective in two complementary senses. 
First, problems should stay near the model's current capability boundary during training~\cite{parashar2025curriculum,chen2025scale}, i.e., remain neither trivial nor unsolvable. When the model mostly sees easy problems, learning saturates; when it mostly sees overly difficult problems, exploration becomes unproductive and the reward becomes sparse.
Second, the training distribution should retain sufficient diversity over time~\cite{li2025learnalignreasoningdataselection}. Even if difficulty is well matched, repeatedly interacting with a narrow task distribution can lead to overfitting to a limited set of patterns, weakening generalization and reducing exploration. 
Note that difficulty variation is not equivalent to diversity. Even if a single environment can generate infinitely many problems by scaling parameters (e.g., longer arrays or larger graphs), it still shares a limited set of templates and failure modes, so learning can plateau once those patterns are mastered.

To this end, we propose \textbf{SCALER} (\textbf{S}ynthetic s\textbf{C}alable \textbf{A}daptive \textbf{L}earning \textbf{E}nvironment for \textbf{R}easoning), a system that combines scalable reasoning environment synthesis with adaptive multi-environment reinforcement learning.
For the first limitation, SCALER provides a way to generate verifiable tasks at scale with controllable difficulty.
We innovatively develop a synthesis pipeline that programmatically converts real-world programming problems into reasoning environments with (i) verifiable interaction via deterministic oracles and unit tests, (ii) controllable difficulty via explicit scale parameters, and (iii) unbounded instance generation within each environment through randomized testcase generation. The pipeline automatically extracts meta-information, validates testcase generators via breadth/depth checks. This enables scaling training beyond finite reasoning datasets or a small set of hand-crafted environments, while retaining strong correctness guarantees.

On the other hand, SCALER designs an adaptive multi-environment RL framework to adapt both instance difficulty and environment selection so that training continues to encounter informative challenges as the model improves.
Specifically, within each environment, an online difficulty controller adjusts scale parameters based on on-policy rollout accuracy to keep sampled instances near a target success rate, thereby tracking the model`s capability frontier and avoiding degenerate ``all-correct'' or ``all-wrong'' regimes. Across environments, an environment curation mechanism maintains an active set of environments and replaces those whose learning signal has saturated (e.g., difficulty no longer increases or the environment becomes consistently trivial/unlearnable). This realizes an environment-level effectiveness. As the model and environments co-evolve, the marginal learning benefit can diminish, so continuously refreshing the active set helps preserve novelty and sustained learning signals.

Our contributions can be summarized as follows:
\begin{itemize}[leftmargin=*, topsep=0pt, partopsep=0pt, parsep=0pt, itemsep=0pt]
  \item We highlight the importance of difficulty and diversity controllable environments for scaling RL post-training in reasoning.
  \item We propose \textbf{SCALER} that combines verifiable, difficulty-controllable environment synthesis with adaptive multi-environment RL.
  \item Extensive experiments demonstrate SCALER yields consistent improvements across diverse reasoning benchmarks and exhibits more sustained training dynamics than other RL baselines under comparable budgets.
\end{itemize}
\section{Related Work}

\label{sec:related-work}

\paragraph{Data-centric RL.}
A data-centric line of work improves the scalability of RL for LLM reasoning by continuously expanding the training distribution with synthetic data~\cite{wu2025synthrlscalingvisualreasoning,setlur2024rlincorrectsyntheticdata} and self-play~\cite{fang2025serlselfplayreinforcementlearning,liang2025swsselfawareweaknessdrivenproblem,wang2025improvingrationalityreasoningprocess,chen2025spcevolvingselfplaycritic}.
On the synthetic-data side, works like Synthetic Data RL~\cite{guo2025syntheticdatarltask} and SWiRL~\cite{goldie2025syntheticdatageneration} generate task-specific supervision from question--answer pairs to multi-step reasoning and tool-use trajectories, and then apply RL on the resulting synthetic corpus.
On the self-play side, recent methods produce interactions and hard cases via adversarial games or role-based play, enabling continual data self-generation for policy improvement~\cite{zhao2025absolute,liu2025spiral}.
However, these approaches are easy to encounter bottlenecks: as the max difficulty of tasks is bounded by the generator agent~\cite{chae2025understandingselfplayllmreasoning}, the data may drift out of sync with the evolving policy, collapsing into instances that are too easy or too hard and thus weakening learning signals in later stages.
In contrast, SCALER  applies difficulty controller and environment curation mechanism on synthesized  verifiable reasoning environment to keep training informative throughout RL.

\paragraph{Difficulty-aware RL.}
To mitigate vanishing learning signals, difficulty-aware RL often employs curriculum learning~\cite{kimi1.5,shi2025efficient,chen2025self} or difficulty scheduling~\cite{wang2025schedulingllmreinforcementlearning,chen2025datacentricsamplecentricenhancingllm,tong2025game}.
Curriculum-based method trains models from easy to hard~\cite{liu2025saturn,parashar2025curriculumreinforcementlearningeasy}, keeping effective learning signals in training process.
However, curricula designed over static dataset faces the challenge of being coarse-grained and hard to design.
This motivates environment-based formulations where difficulty of each environment can be monitored and adjusted online~\cite{guo2025genenvdifficultyalignedcoevolutionllm}.
Reasoning Gym~\cite{stojanovski2025reasoning} offers a suite of procedurally generated, verifiable reasoning environments with tunable difficulty, while RLVE~\cite{zeng2025rlve} further adapts the difficulty distribution online as the policy improves.
Despite enabling difficulty tracking and adaptation, existing environment suites are largely hand-engineered and limited in environment diversity at scale.
SCALER addresses this by automatically synthesizing a large and diverse set of reasoning environments with verifiable oracles and controllable difficulty, supporting sustained multi-environment RL.

\section{\textbf{SCALER}}
\label{sec:method}
\begin{figure*}[t]
    \centering
    \includegraphics[width=\linewidth]{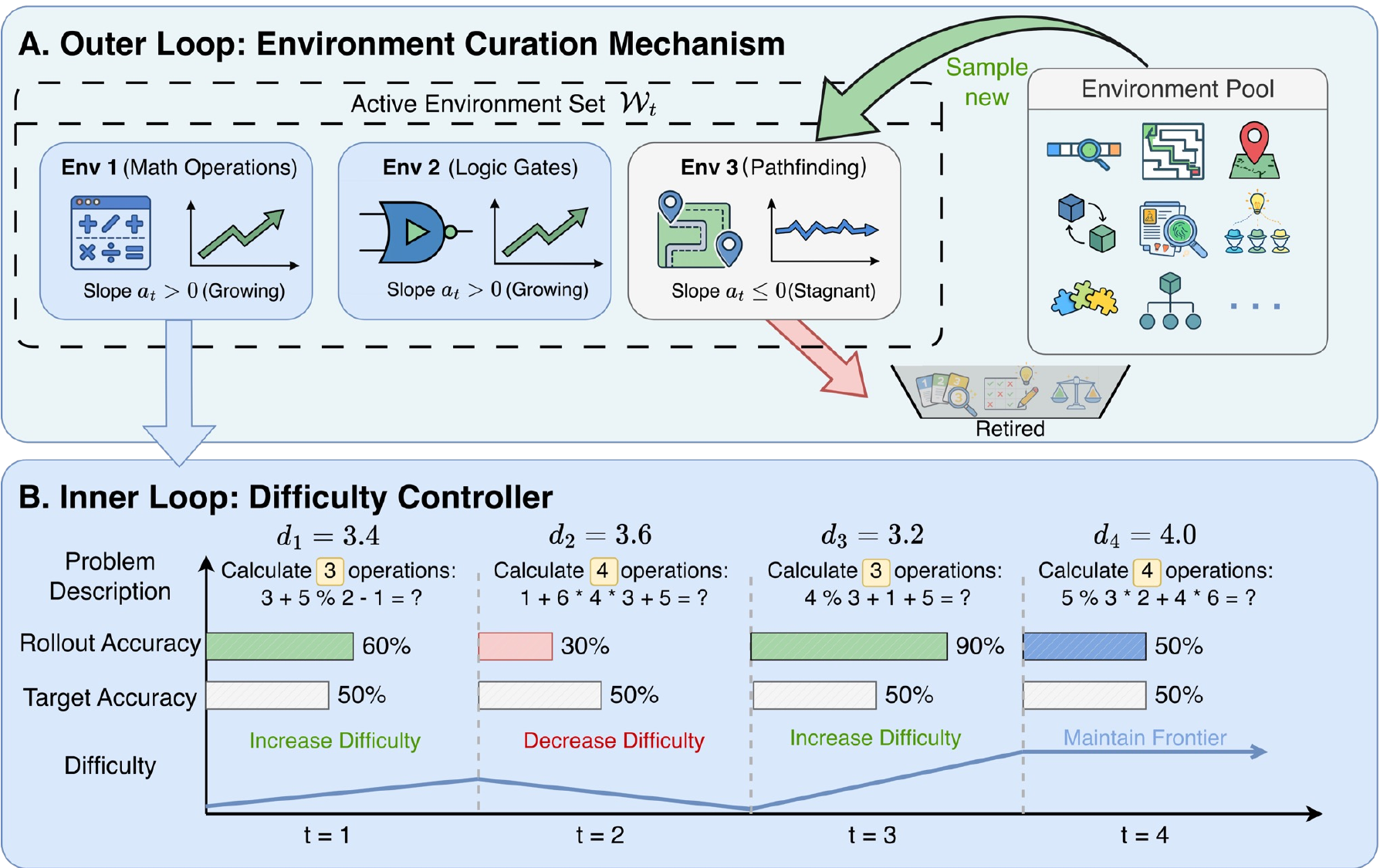}
    \caption{An illustrative training example of \textbf{SCALER}. The model interacts with a set of active environments: instances are used for training, online accuracy updates the in-environment difficulty controller, and environments whose learning signal saturates are retired and replaced by new environments via the curation mechanism.}
    \label{fig:rl_example}
\end{figure*}

To enable the model to explore, adapt, and generalize in multiple dynamic environments, we design \textbf{SCALER} with two components: (i) an adaptive multi-environment training framework that learns through interaction with a set of verifiable environments, and (ii) a systematic synthesis pipeline that converts programming problems into difficulty-controllable reasoning environments.  
Specifically, we first introduce the learning strategy over environments, including in-environment difficulty controller and the environment curation mechanism (\S\ref{sec:train_framework}), then describe the environment synthesis process from programming problems (\S\ref{sec:systhesize_pipeline}).

\begin{algorithm*}
\caption{SCALER Training Framework}
\label{alg:scaler}
\begin{algorithmic}[1]
\REQUIRE Policy $\pi_\theta$, Environment Pool $\mathcal{E}_{pool}$, Target accuracy $\tau$, Step size $\beta$, Active set size $M$
\STATE \textbf{Initialize:} Sample initial active set $\mathcal{W} \leftarrow \{e_1, \dots, e_M\} \subset \mathcal{E}_{pool}$,Difficulty $d \leftarrow d_{min}$ for all $e \in \mathcal{W}$
\FOR{step $t = 1, 2, \dots, T_{max}$}
    \STATE $\mathcal{D}_{batch} \leftarrow \emptyset$
    \FOR{each environment $e \in \mathcal{W}$}
        \STATE Sample instances $x$ from $e$ based on difficulty $d$ 
        \STATE Collect trajectories using $\pi_\theta$ and calculate accuracy $\mathrm{acc}_t$
        \STATE $\mathcal{D}_{batch} \leftarrow \mathcal{D}_{batch} \cup \text{Trajectories}$
        
        \STATE Update difficulty $d$ for this environment \hfill $\triangleright$ Eq.~\ref{eq:diff-update}
        
    \ENDFOR
    
    \STATE Update $\pi_\theta$ with $\mathcal{D}_{batch}$ 
    
    \FOR{each environment $e \in \mathcal{W}$}
    
        \STATE Calculate slope $a^e_t$ over last $K_{slope}$ steps  \hfill $\triangleright$ Eq.~\ref{eq:slope_lr}
        
        \IF{$a^e_t \le 0 \lor \mathrm{acc} = 0 \text{ for } K_{zero} \text{ steps} \lor d = D^e \text{ for } K_{sat} \text{ steps}$}
            \STATE Retire $e$: $\mathcal{W} \leftarrow \mathcal{W} \setminus \{e\}$; $\mathcal{E}_{pool} \leftarrow \mathcal{E}_{pool} \cup \{e\}$
            \STATE Sample new $e_{new} \sim \mathcal{E}_{pool}$
            \STATE $\mathcal{W} \leftarrow \mathcal{W} \cup \{e_{new}\}$
        \ENDIF
    \ENDFOR
\ENDFOR

\end{algorithmic}
\end{algorithm*}

\subsection{Multi-environment Training Framework}
\label{sec:train_framework}
 
Following~\cite{zeng2025rlve}, we can formally define  an environment as a tuple $E = (I,P, R)$, representing a context-conditioned, parameterized problem space where individual instances evolve through controllable parameters.
Here $I$ is an input template, $P$ is a problem generator, and $R$ is a verifier.
In a given environment, $P$ generates an unbounded set of problems following the same input template $I$, while $R$ provides the corresponding ground truth for all generated problems.
For example, as shown in the math-operations environment in Figure~\ref{fig:rl_example}, all problems share a common contextual description $I$, \textit{calculate $N$ operations}, problem generator  $P$ controls the difficulty by the scale parameter $N$ and synthesizes the array, and the programming solution serves as $R$. 
More illustrative cases are provided in Appendix~\ref{app:environment_cases}.

To fully unleash the potential of these  difficulty-controllable reasoning environments, we adopt two complementary learning strategies:
(i) Difficulty Controller. In environments with explicit verifiability, we dynamically adjust the environment's scale parameters according to the online accuracy, which provides continuous and stable learning signals.
(ii) Environment Curation Mechanism. For general environments, we introduce an environment curation mechanism within which the model is trained exclusively on the current active set of environments. As the model and the environment gradually co-adapt and the learning signal diminishes in marginal utility, the current environment is retired and replaced with a newly sampled one.

The detailed implementation of SCALER is provided in Algorithm~\ref{alg:scaler}.

\subsubsection{Difficulty Controller}
\label{sec:difficulty-control}


A key goal of SCALER is to maintain training at the agent's capability frontier by continuously sampling instances near the boundary of what the current policy can solve. To achieve this, difficulty is dynamically adjusted within each environment based on current accuracy.

Specifically, the difficulty of the instances is characterized by the array length or the number of edges in the graph and we discretize the difficulty into distinct difficulty levels. 
Let $\mathrm{acc}_t \in [0,1]$ denote the average accuracy over $k$ sampled instances at step $t$ in a given environment, and let $\tau \in [0,1]$ denote the target accuracy.
The continuous difficulty score $d_t \in \mathbb{R}$, initializing $d_0 = 0$, evolves according to:
\begin{equation}
d_{t+1} = \mathrm{clip}(d_t + \beta \cdot (\mathrm{acc}_t - \tau),\ 0,\ D),
\label{eq:diff-update}
\end{equation}
where $\beta > 0$ controls the adaptation rate and $D \in \mathbb{N}$ is the maximum discrete difficulty level in the environment.
This update rule increases difficulty when $\mathrm{acc}_t > \tau$ and decreases it otherwise, thereby maintaining the training distribution near the agent's capability boundary.



For the next sampling step, while $d_{t+1}$ is real-valued, the environment can only generate $k$ instances at integer-valued levels in $\{0, 1, \dots, D\}$. To approximate $d_{t+1} \in \mathbb{R}$ under this integer constraint, we construct a multiset of $k$ integers whose mean closely matches $d_{t+1}$. Specifically, let $\ell = \lfloor d_{t+1} \rfloor$ and define $h = \mathrm{round}\!\big(k(d_{t+1} - \ell)\big)$. We assign $h$ instances the value $\ell + 1$ and the remaining $k - h$ instances the value $\ell$, thereby balancing rounding up and rounding down; for example, when $d_{t+1} = 2.3$ and $k = 10$, this yields three instances at level $3$ and seven at level $2$, whose average equals $2.3$.


The proposed controller naturally tracks model competence.
As the model improves, online accuracy rises above the target and difficulty is pushed upward to stay near the performance boundary.
Conversely, if the model forgets or the policy distribution shifts, accuracy drops and the controller reduces difficulty, recovering usable learning signals without manual retuning.

\subsubsection{Environment Curation Mechanism}
\label{sec:curation-mechanism}

To keep sufficient diversity during training and avoid a "tourist" learning pattern where the model randomly explores a variety of environments without sustained progress, SCALER introduces the environment curation mechanism which maintains an active set of environments and restricts training to this dynamic set, continuously refreshing it to avoid spending budget on environments that no longer provide meaningful learning signals.
 
Let $\mathcal{W}_t = \{e_1,\dots,e_{|\mathcal{W}|}\}$ denote the active environment set at training step $t$.
At each step, instances are sampled only from environments in $\mathcal{W}_t$ and used for policy optimization.
Environments are periodically evaluated at every step and may be retired from the set and replaced by newly synthesized environments.

For each environment $e$, let $d_t^{e}$ denote its  difficulty at step $t$.
To detect stalled progress, we argue that when the recent difficulty-step slope is less or equal to zero, the learning on this environment is saturated and should be retired.
Specifically, we estimate the slope by fitting a least-squares line over the last $K_{slope}$ steps, defining the index set $T=\{t-K_{slope}+1,\dots,t\}$ and fitting
\begin{equation}
d_i^{e} \approx a^{e}_t \cdot i + b^{e}_t,\  i \in T,
\end{equation}
where the slope $a^{e}_t$ is given in closed form by
\begin{equation}
a^{e}_t \;=\;
\frac{\sum_{i\in T} (i-\bar{i})(d_i^{e}-\bar{d}^{e})}
{\sum_{i\in T} (i-\bar{i})^2},
\label{eq:slope_lr}
\end{equation}
with $\bar{i}=\frac{1}{|T|}\sum_{i\in T} i$ and $\bar{d}^{e}=\frac{1}{|T|}\sum_{i\in T} d_i^{e}$.
An environment is retired if $a^{e}_t \le 0$, indicating that difficulty is temporarily saturated  in that environment.

To accelerate retirement decisions, two additional heuristics are applied.
Let $\mathrm{acc}_t^{e} \in [0,1]$ denote the accuracy at step $t$ and let $D^{e}$ be the maximum difficulty of environment $e$.
An environment is retired early if it is unlearnable, i.e., $\mathrm{acc}_t^{e}=0$ for $K_{zero}$ consecutive steps, or saturated, i.e., $d^{e}_t=D^{e}$ for $K_{sat}$ consecutive steps.

When an environment is retired, it is replaced by a newly synthesized environment sampled from the environment pool, keeping the set size fixed.
Such retirement is temporary, as it targets situations where the model has reached a plateau in learning within the current environment, rendering further progress unproductive. However, as training continues and the model's capability boundary improves, there remains potential for continued learning within these environments. Therefore, retired environments are reintegrated into the environment pool and will be resampled in future iterations, offering renewed opportunities for the model to explore and adapt.

In general, the environment curation mechanism supports sustained learning within multiple environments while preventing training from stagnating on environments that have become either trivial or unproductive.

\subsection{From Code Generation to Reasoning Environments}
\label{sec:systhesize_pipeline}
Our multi-environment training framework continuously tracks the model's capability frontier while maintaining data diversity and freshness. 
This requires a steady supply of environments that are difficulty-controllable and automatically verifiable. 
Since real-world programming problems provide problem descriptions as the input template $I$, codes as the verifier $R$ for testcases at varying difficulty levels, we therefore introduce an environment synthesis pipeline that converts these problems into SCALER environments.

Accordingly, our core objective is to synthesize the problem generator $P$, which can be organized into three components:
\begin{itemize}[leftmargin=*, topsep=0pt, partopsep=0pt, parsep=0pt, itemsep=0pt]
    \item \textbf{Meta information extraction} (\S\ref{sec:extract-metadata}).
    Ensures rewardable and unambiguous supervision by selecting problems with well-defined, verifiable outputs, mitigating output-format mismatch and reward hacking.
    \item \textbf{Testcase generation and verification} (\S\ref{sec:testcase-gen}).
    Provides a stream of diverse, valid, and automatically verifiable instances, addressing the mismatch between static problem statements and stochastic environment inputs.
    \item \textbf{Heuristic difficulty calibration} (\S\ref{sec:difficulty-setting}).
    Defines an executable difficulty range under context and runtime budgets, addressing the scale mismatch between real-world inputs and agentic prompting.
\end{itemize}

After filtering proper programming problems with extracted meta information, we can generate testcases of specific difficulty, eventually constructing difficulty-controllable SCALER environments.

\subsubsection{Extracting Meta Information}
\label{sec:extract-metadata}



To offer high-quality programming problems for reasoning environment synthesis and mitigate the risk of reward hacking, we adopt a prompt-based method to extract key meta-information from candidate problems, followed by a rule-based filtering process. 
The complete extraction prompt is provided in Appendix~\ref{app:extract_prompt}.

For each programming problem \(p\), we extract a metadata tuple \(\texttt{meta}(p)\) comprising (i) scale parameters that characterize complexity-related constraint variables, and (ii) output requirements that describe the output type and whether the correct output is unique. 
Accordingly, we discard ill-formed problems and retain only those with a unique output whose type lies in \(\{\texttt{number}, \texttt{array}, \texttt{string}\}\).

\subsubsection{Generating and Verifying Testcases}
\label{sec:testcase-gen}

To sample different instances with fixed extracted scale parameters from one specific environment, we construct a testcase generator agent that takes a target scale parameter configuration and produces an input whose content is randomized but conforms to the original problem specification.
We provide the full prompting details in Appendix~\ref{app:testcase_prompt}.
Since the environment relies on the synthetic agent, we validate generator functions with two complementary checks.

\paragraph{Breadth check.}
We sample diverse scale parameters and corresponding inputs , following~\cite{fu2025klearcodetestscalabletestcase}, each instance is evaluated by multiple independent ground-truth solutions.
Consistency across solutions is required, which simultaneously (i) detects malformed generator functions and (ii) re-validates the assumption that the output is unique.

\paragraph{Depth check.}
For a fixed scale parameter configuration, we call generator function multiple times and compare the resulting ground-truth outputs across generated instances.
Based on the numbers and max size requirement of output clustering, we enforce sufficient diversity to prevent the training process from overfitting to a narrow pattern distribution or exploiting reward hacking.

\subsubsection{Calibrating Heuristic Difficulty Levels}
\label{sec:difficulty-setting}
As inputs for real-world programming problems can be in the millions, which is not suitable for the prompt used in agentic training, we aim to re-define the difficulty range for each environment that is both scalable and executable.
Apparently, the effective maximum difficulty is bounded by two practical constraints: (i) the maximum prompt length accepted by the policy model, and (ii) the execution time limit enforced by the original problem setting.

To estimate the largest feasible scale parameter configuration $\mathbf{s}_{\max}$ under the prompt-length constraint, a binary search is performed over a single global scale factor. 
Feasibility is determined by whether the testcase fits within the context window while respecting the fixed execution time limit.

Given $\mathbf{s}_{\min}$ and $\mathbf{s}_{\max}$, the environment defines a finite set of difficulty levels by discretizing the scaling range.
To keep the number of levels within a reasonable budget, the discretization strategy depends on the span of each parameter: when the range is small, levels follow an arithmetic progression; when the range is large, levels follow a geometric progression.
This heuristic yields a compact but expressive difficulty ladder, enabling smooth scheduling while preserving meaningful granularity across scales.



    
    


\section{Experiments}



\label{sec:experiments}

SCALER addresses two key challenges: (i) providing a synthesis pipeline that systematically generates environments with infinite data generation capabilities, and (ii) developing a multi-environment training framework, which enables sustaining model improvement through adaptive learning across diverse environments.
To validate the efficiency of our approach, experiments are organized around four research questions~(RQs):
\begin{itemize}[leftmargin=*, topsep=0pt, partopsep=0pt, parsep=0pt, itemsep=0pt]
    \item RQ1: How does SCALER compare with RL training baselines under comparable training budgets? 
     \item RQ2: How does performance vary as the number of environments increases?
     \item RQ3: Are all components of SCALER necessary for their gains?
     \item RQ4: How sensitive is SCALER to hyperparameters?
\end{itemize}

\subsection{Experimental Setup}
\label{sec:exp-setup}
\paragraph{Synthesis Pipeline Settings.}
We use CodeContests~\cite{codecontests} as the seed dataset and GLM-4.6~\cite{zeng2025glm} as the data-synthesis agent. For code execution, we use SandboxFusion~\cite{bytedanceseedfoundationcodeteam2025fullstackbenchevaluatingllms}. After filter, we obtain the subset consisting of 4973 programming problems and synthesize 2739 SCALER environments.
Specifically, we offer environment cases in Appendix~\ref{app:environment_cases}.

\paragraph{Training setup.}
All experiments start from Qwen3~\cite{qwen3technicalreport}series, and we use Qwen3-1.7B-base and Qwen3-4B-base as the policy model.
For the environment curation mechanism, we set $K_{slope}=10$ and $K_{zero}=K_{sat}=5$.
The training batch size equals the environment set size: at each optimization step, one problem is sampled from each of the 64 environments according to the environment-specific difficulty controller, resulting in 64 prompts per step.
Reinforcement learning is performed with GRPO~\cite{shao2024deepseekmath}.
For each prompt, $n_{\text{resp}}{=}8$ responses are sampled.
Training-time decoding uses temperature $T{=}1.0$ and ${top\_p}{=}1.0$.
The prompt length budget is capped at 4096 tokens and the response length budget at 8192 tokens.
Additional hyperparameters and infrastructure details are deferred to Appendix~\ref{app:train_details}, and we provide computational cost in Appendix~\ref{app:computational_cost}.

\paragraph{Evaluation protocol.}
Performance is evaluated on five benchmarks: AIME24~\cite{aime2024}, AMC23~\cite{amc2023}, MATH-500~\cite{math500}, MMLU-Pro~\cite{mmlu_pro}, and BBEH~\cite{bbeh}.
Results are reported as avg@16 for AIME24, AMC23 and MMLU-Pro, avg@1 for the remaining benchmarks.
Unless otherwise specified, decoding uses temperature $T=0.6$ and ${top\_p}=0.95$.

\subsection{RQ1: SCALER outperforms both dataset-based and environment-based baselines}
\label{sec:rq1}

\begin{table*}[t]
\centering
\small
\setlength{\tabcolsep}{9pt}
\renewcommand{\arraystretch}{1.15}
\begin{tabular}{lcccccc}
\toprule
Method & MATH-500 & AMC23 & AIME24 & MMLU-Pro & BBEH & AVG \\
\midrule
\multicolumn{7}{l}{\textit{Qwen3-1.7B-Base}} \\
\midrule
Base            & 59.60 & 29.21 &  3.33 & 33.30 &  3.26 & 25.74 \\
DeepMath-103K   & 73.60 & 47.97 & 14.58 & 49.64 &  9.56 & 39.07 \\
MATH-7.5k       & 75.60 & \textbf{50.78} & \textbf{15.20} & 46.78 &  6.08 & 38.89 \\
RLVE            & 70.80 & 47.03 & 12.91 & 46.96 & 11.30 & 37.80 \\
\rowcolor{gray!15} \textbf{SCALER (Ours)} & \textbf{75.80} & 49.53 & 12.91 & \textbf{50.89} & \textbf{11.74} & \textbf{40.18} \\
\midrule
\multicolumn{7}{l}{\textit{Qwen3-4B-Base}} \\
\midrule
Base            & 66.40 & 44.70 &  8.75 & 51.60 &  8.10 & 35.91 \\
DeepMath-103K   & \textbf{86.60} & 65.60 & 22.29 & 68.03 & 12.82 & 51.08 \\
MATH-7.5k       & 86.20 & 70.63 & 24.16 & 69.19 & 10.00 & 52.04 \\
RLVE            & 84.80 & 70.94 & 25.42 & 69.91 & \textbf{16.52} & 53.52 \\
\rowcolor{gray!15} \textbf{SCALER (Ours)} & 84.40 & \textbf{75.00} & \textbf{27.29} & \textbf{70.00} & 14.56 & \textbf{54.25} \\
\bottomrule
\end{tabular}
\caption{Performance comparison of RL baselines and SCALER on five reasoning benchmarks. AVG is the unweighted mean over the five benchmarks. The highest performance is \textbf{bolded}.}
\label{tab:rq1-main-results}
\end{table*}

\begin{figure*}[t]
    \centering
    \includegraphics[width=0.49\linewidth]{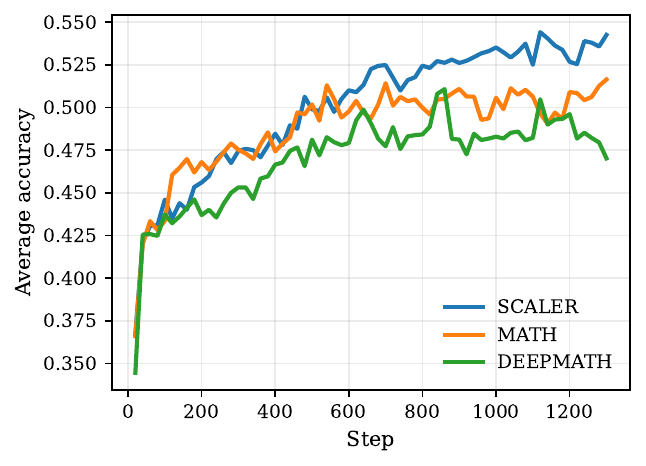}
    \hfill
    \includegraphics[width=0.49\linewidth]{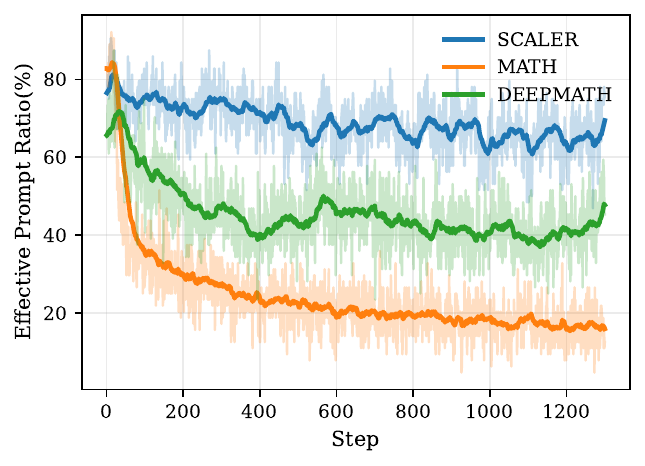}
    \caption{
    Left: average performance across the five evaluation benchmarks during Qwen3-4B-base training, comparing dataset-based baselines (MATH, DeepMath) and SCALER.
    Right: effective sampling statistics under SCALER, indicating that most sampled instances remain near the model's capability boundary.
    }
    \label{fig:rq1_perf_and_efficiency}
\end{figure*}
A common paradigm for improving reasoning is to scale up training on static datasets, such as curated corpora with verified solutions (e.g., MATH~\cite{hendrycks2021measuringmathematicalproblemsolving}) or collections augmented by teacher-generated answers (e.g., DeepMath~\cite{he2025deepmath}). 
However, static corpora provide limited learning signals, restricting an agent's ability to adapt to new environments and generalize beyond seen distributions. 
While interactive environment-based learning, such as RLVE~\cite{zeng2025rlve}, has emerged to address this limitation, its effectiveness can still be bounded by environment diversity. 
To comprehensively evaluate SCALER, this RQ compares it against both paradigms under comparable training budgets.

As shown in Table~\ref{tab:rq1-main-results}, while both the dataset-based baselines and the environment-based baseline  substantially improve upon the base models, SCALER consistently achieves the highest overall average. 
Specifically, SCALER not only significantly surpasses the static dataset baselines—confirming the necessity of interactive learning—but also outperforms RLVE, demonstrating the superiority of our specific environment-based design. 
Furthermore, SCALER delivers consistent improvements across all five evaluations, including MMLU-Pro and BBEH, suggesting stronger transferability beyond narrowly curated math domains.

The training dynamics in Figure~\ref{fig:rq1_perf_and_efficiency} provide additional evidence for this gap.
Figure~\ref{fig:rq1_perf_and_efficiency} (left) shows that SCALER not only reaches a higher level, but also exhibits a more sustained improvement trend: the aggregated evaluation performance continues to rise for more than 1{,}000 training steps, whereas dataset-based baselines plateau earlier.
Importantly, this behavior is supported by boundary-focused sampling: Figure~\ref{fig:rq1_perf_and_efficiency} (right) indicates that SCALER keeps a higher effective sampling rate than dataset-based baselines, where most prompts remain neither trivial nor intractable.
By continuously providing instances near the model's current capability boundary from SCALER environments, SCALER sustains informative reward during RL, mitigates premature saturation, and enables longer-horizon performance gains.

\subsection{RQ2: Scaling environment size leads to incremental performance gains}
\label{sec:rq2}

\begin{figure}[t]
    \centering
    \includegraphics[width=0.49\linewidth]{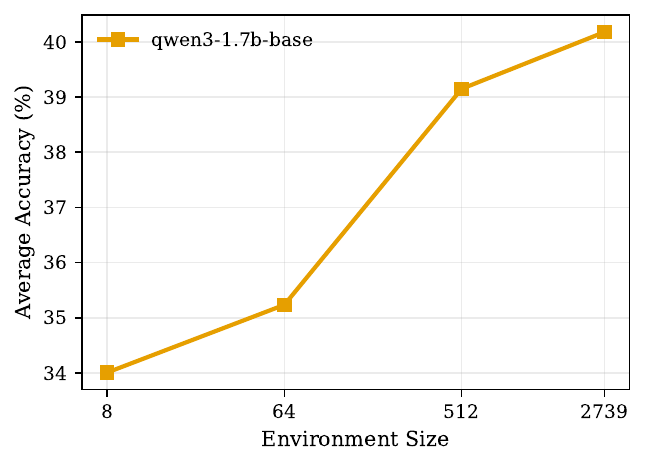}
    \hfill
    \includegraphics[width=0.49\linewidth]{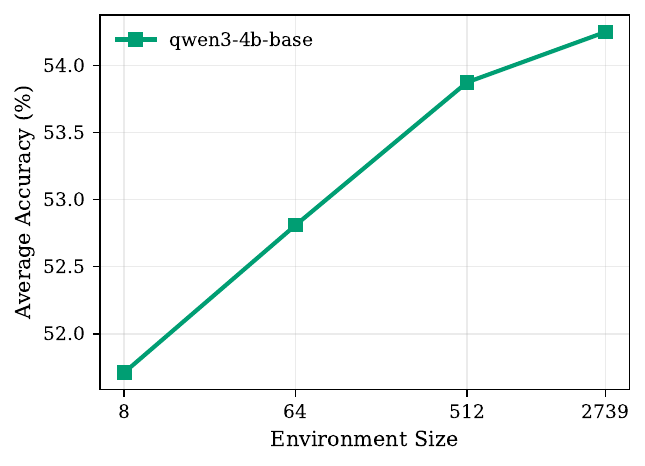}
    \caption{Accuracy improvements for both Qwen3-4B-base and Qwen3-1.7B-base as environment size increases. Both models show a consistent increase in performance with larger environment sizes.}
    \label{fig:rq2-env-scaling}
\end{figure}

SCALER introduces the automatic environment synthesis pipeline which significantly reduces labor costs.
This RQ examines how model performance changes as the number of environments increases, with results summarized in Figure~\ref{fig:rq2-env-scaling}.

We conduct the experiment by randomly sample 8, 64 and 512 environments from 2739 SCALER environments, where the larger size of environments always contains the smaller size one.
As shown in Figure~\ref{fig:rq2-env-scaling}, increasing the number of environments from 8 to 2739 leads to incremental performance improvements. The model benefits from encountering a greater diversity of tasks, which allows it to maintain consistent learning progress.
With exposure to a wider variety of challenges and environments facilitating the development of broader reasoning skills, scaling the number of environments enables the model to continuously adapt and enhance its reasoning capabilities.

\begin{figure}[t]
    \centering
    \includegraphics[width=0.5\linewidth]{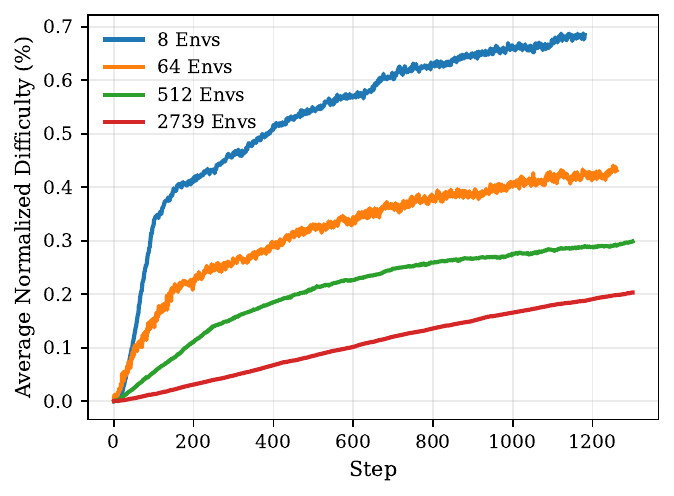}
    \caption{Training dynamics of Qwen3-4B-Base with different numbers of environments. The plot shows that even with smaller environments, the model continues to learn with increasing difficulty levels.}
    \label{fig:rq2-env-dynamics}
\end{figure}

It is worth noting that even with a smaller number of environments, the model continues to engage in ongoing exploration, with increasing difficulty levels within each individual environment, as illustrated in Figure~\ref{fig:rq2-env-dynamics}. 
The key insight is that scaling the number of environments with fixed training budgets, inherently involves a trade-off between task difficulty and diversity. Striking the right balance is critical: focusing solely on increasing difficulty may result in limited improvements, while excessive diversity without a proper difficulty controller, i.e. DeepMath dataset with 103k samples, does not necessarily lead to optimal performance.


\subsection{RQ3: All components of \textsc{SCALER} are necessary for its gains}
\label{sec:rq3}
\begin{table*}[h]
\centering
\small
\setlength{\tabcolsep}{5pt}
\renewcommand{\arraystretch}{1.1}
\begin{tabular}{lcccccc}
\toprule
$K_{slope}$ & \textbf{MATH-500} & \textbf{AMC23} & \textbf{AIME24} & \textbf{MMLU-Pro} & \textbf{BBEH} & \textbf{AVG} \\
\midrule
5  & 73.00 & \textbf{52.34} & 13.54 & 50.71 & 10.22 & 39.96 \\
\textbf{10} & \textbf{75.80} & 49.53 & 12.91 & \textbf{50.89} & \textbf{11.74} & \textbf{40.18} \\
20 & 75.60 & 47.03 & \textbf{14.79} & 49.46 & 10.87 & 39.55 \\
\bottomrule
\end{tabular}
\caption{Sensitivity to $K_{slope}$ on Qwen3-1.7B.}
\label{tab:k_slope}
\end{table*}

\begin{table*}[h]
\centering
\small
\setlength{\tabcolsep}{5pt}
\renewcommand{\arraystretch}{1.1}
\begin{tabular}{lcccccc}
\toprule
$K_{zero} = K_{sat}$ & \textbf{MATH-500} & \textbf{AMC23} & \textbf{AIME24} & \textbf{MMLU-Pro} & \textbf{BBEH} & \textbf{AVG} \\
\midrule
1  & 74.40 & 46.09 & \textbf{15.00} & 49.64 &  9.35 & 38.89 \\
\textbf{5}  & \textbf{75.80} & \textbf{49.53} & 12.91 & \textbf{50.89} & \textbf{11.74} & \textbf{40.18} \\
10 & 72.40 & 45.94 & 11.25 & 49.29 & \textbf{11.74} & 38.12 \\
\bottomrule
\end{tabular}
\caption{Sensitivity to $K_{zero}$ and $K_{sat}$ on Qwen3-1.7B.}
\label{tab:k_sat}
\end{table*}

\begin{figure}[t]
    \centering
    \includegraphics[width=0.5\linewidth]{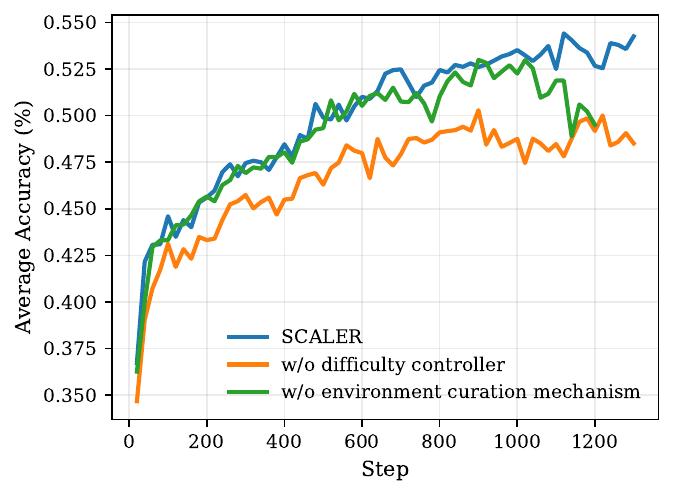}
    \caption{Ablation study of \textbf{SCALER} on Qwen3-4B-Base. Validation accuracy over training steps for the full system and two variants that remove difficulty controller or the environment curation mechanism. Both components contribute to stronger and more sustained performance improvements.}

    \label{fig:rq3-ablation}
\end{figure}

This experiment studies whether \textsc{SCALER}'s improvements require its two core components in multi-environment training framework: adaptive difficulty controller and environment curation mechanism.

Two ablations based on Qwen3-4B-base are compared against the full SCALER system in Figure~\ref{fig:rq3-ablation}.
Removing the curation mechanism disables environment replacement, forcing training to randomly sample every environment.
To model the ablation of difficulty controller, we replace the boundary tracking with random difficulty sampling, selecting 5 instances per environment. This modification reduces the consistency of the supervision signal, as the environment selection is no longer guided by a difficulty-based approach.

Both ablations lead to lower performance than the full system, indicating that difficulty controller and the curation mechanism each make a material contribution.
Qualitatively, difficulty controller prevents training from drifting into regimes that are too easy or too hard, while the curation mechanism promotes sustained learning within learnable environments and mitigates shallow adaptation.



\subsection{RQ4: SCALER is not sensitive to hyperparameters}
\label{app:hyperparameter}

To ensure the robustness of our approach, we conducted a comprehensive sweep of hyperparameters, specifically focusing on $K_{slope}$, $K_{zero}$, and $K_{sat}$. Overall, we found that SCALER is not highly sensitive to these hyperparameter settings. 

To test the sensitivity of SCALER to the history-window parameters, we conducted hyperparameter experiments on the Qwen3-1.7B model. Table~\ref{tab:k_slope} and Table~\ref{tab:k_sat} present the performance of SCALER across different values of $K_{slope}$ and $K_{zero}/K_{sat}$. The results demonstrate that SCALER remains relatively stable across these settings.
The rationale behind the performance variations can be explained as follows:
\begin{itemize}
    \item \textbf{Impact of $K_{zero}$ and $K_{sat}$:} Setting these thresholds too small retires environments too aggressively. This may prematurely discard environments that are temporarily in a slump but could otherwise recover. Conversely, setting them too large keeps unsuitable environments active for too long, wasting compute resources on low-value training.
    \item \textbf{Impact of $K_{slope}$:} A small value makes the algorithm overly sensitive to short-term noise. However, a large value dilutes the slope signal with stale history, causing the system to react too slowly to real performance degradation.
\end{itemize}

Based on these empirical observations and trade-offs, we heuristically set $K_{slope} = 10$ and $K_{zero} = K_{sat} = 5$ as our default configuration, which effectively strikes an optimal balance between responsiveness and stability.

\section{Conclusion}
In this paper, we propose a novel approach SCALER containing two key components: (i) reasoning environment synthesis pipeline, (ii) multi-environment training framework. 
Through our synthesis pipeline for generating a variety of difficulty-controllable environments , our approach offers a platform for the research community to explore the impact of properties of environments on RL training.
Moreover, the multi-environment training framework ensures the difficulty of instances consistently matches the model's capabilities and maintains diversity and freshness, jointly preserving the \textbf{continuously effective} reward signals to improve model capabilities in long horizon.

\section*{Limitations}
While our work provides a comprehensive analysis of SCALER's ability to improve model performance through dynamic environment synthesis and multi-environment training, there are several limitations that warrant further exploration:
\begin{itemize}[leftmargin=*, topsep=0pt, partopsep=0pt, parsep=0pt, itemsep=0pt]
\item \textbf{Exploration of Environment Internal Attributes}: Our study primarily focuses on scaling environment size and its components. However, the internal characteristics of environments, such as the richness of context, intrinsic difficulty, and other environment-specific properties, have yet to be thoroughly investigated. Future research should examine how these factors influence model performance, especially in the context of dynamic difficulty adjustment and environment transitions during training.
\item \textbf{Limited Scope of Environments}: Our study is based on 2739 SCALER environments, and while this is a larger number compared to hand-crafted ones, the scaling still hasn't been fully explored in our experiments.
Further research is needed to explore scaling laws related to environment size, model size, and computational resources, and to understand how these scaling factors impact both model training efficiency and performance.
\end{itemize}

\section*{Acknowledgments}
This project was supported by the National Natural Science Foundation of China (NSFC) under Grant No. 62576102.

\appendix
\clearpage
\section{Prompt Design}
\subsection{Extract Meta Information Prompt}
\label{app:extract_prompt}

\begin{lstlisting}[breaklines=true,caption={Prompt used for extracting meta information.}]
You are given as input the full statement of a single algorithmic problem. Your job is to **emit one JSON code block** that captures:
1. All numeric *scale parameters* that are relevant to the **time complexity** of typical solutions.
2. The **type** of the required output.
3. Whether the required output is **unique**, whether the problem has exactly one correct output for each valid input.
## 1. Scale parameters
A *scale parameter* is any integer quantity that bounds:
- The number of items, elements, or positions (n= number of elements, m = number of edges, q = number of queries).
- The size (length) of a grid, string, or sequence (e.g. length up to 2e5, grid up to 1000 * 1000).
- The size of a state space or iteration space that an algorithm must explicitly handle.
### What to INCLUDE as scale parameters
Include a parameter only if **all** of the following are true:
- It directly bounds the size or count of something that is iterated over, e.g.:
  - number of elements / vertices / edges / queries (n, m, q, etc.),
  - length of a string or array,
  - rows / columns of a grid.
- The bounds appear explicitly in the statement, usually in the Input / Constraints section, like:
  - 1 <= n <= 2*10^5
  - 0 <= m <= 2*10^5
- You can clearly identify the parameter name (e.g., n, m, q, k, N, etc.).
### What to EXCLUDE from scale parameters
- Number of test cases / groups: **never** include t, T, or similar when it means "number of test cases".
- Pure value ranges for single items that do **not** change the input size, e.g.:
  - -10^9 <= a_i <= 10^9 when a_i is just the value of an element.
  - Coordinate or weight ranges that are not used as sizes of arrays/grids.
- Any quantity that only affects output format or precision.
### Representation of scale parameters
In the JSON, represent scale parameters under the key "scale_params" as:
<scale_params example>
## 2. Output type classification
You must classify the type of the required output into exactly one of the following strings:
- "string": The required output is a single string or a small number of strings.
- "number": The required output is a single numeric value (integer, real, etc.), e.g. "print one integer as the answer".
- "array": The required output is a one-dimensional sequence (list) of values, e.g. an array of integers, a permutation, a sequence of answers for each query when printed as space-separated numbers or in multiple lines.
- "graph": The required output is a graph structure, such as a set of edges, tree description, adjacency list, or any structure where the output itself is naturally a graph.
- "matrix": The required output is a 2D grid or matrix.
- "bool": The required output is logically a boolean answer, e.g. "YES/NO", "True/False", "Alice/Bob", etc.
- "others": The required output is a complex or mixed structure that does not fit clearly into any of the above categories.
## 3. Output uniqueness
You must also decide whether the required output is unique for each valid input.
Define "is_output_unique" as:
- true if, for any fixed valid input, there is exactly one correct output that satisfies the problem statement.
- false if the statement allows multiple different outputs to be accepted as correct for the same input.
## JSON Output Specification
You must produce exactly one JSON object in a fenced JSON code block.
The JSON must have the following top-level keys:
- "scale_params": an object mapping parameter names to {{ "min": <int>, "max": <int> }}.
- "output_type": one of "string", "number", "array", "graph", "matrix", "bool", "others".
- "is_output_unique": a boolean.
## Example 1: <example 1>
## Example 2: <example 2>
## Example 3: <example 3>
## Final instruction
Now, read the provided problem statement and output the single JSON code block accordingly.
{problem}
\end{lstlisting}
\subsection{Generate Test Case Prompt}

\label{app:testcase_prompt}
\begin{lstlisting}[breaklines=true,caption={Prompt used for constructing generate\_testcase function.}]
You are given as input a single *algorithmic problem statement* (like those from programming contests). Your job is to **emit one Python code block** that defines a *test-case generator* function for this problem.
The generator must produce **exactly one** valid test case per call, parameterized only by the numeric scale values provided via a JSON object.
## Input
You will be given:
1. A raw problem statement in natural language that fully specifies:
   - the input format,
   - the constraints,
   - and the meaning of each variable.
2. An example json_obj instance:
   - This is only an example to clarify field names and typical ranges.
   - Your code must work for any valid json_obj that matches the described schema.
## Required Python output (emit exactly one Python code block)
You must output a Python code block that defines **one single function** with the following signature:
```python
def generate_testcase(json_obj: dict) -> tuple[str, dict]:
    """
    Generate a test case based on the given json_obj.
    Parameters:
    - json_obj (dict): The input JSON object containing problem parameters.
    Returns:
    - tuple[str, dict]: A tuple containing:
      - The first element is a string representing the test case in input format.
      - The second element is a dictionary representing the same test case.
    """
    ...
```
### Return value
- Your function must return both the string and the dictionary representation of the test case in a tuple. The first element of the tuple should be the string format, and the second element should be the dictionary format.
    - output_str:
        - A single string that is a valid input for the problem according to the Input section, representing exactly one logical test case.
        - If the problem statement defines a format with multiple test cases controlled by an integer T in the input, you must set T = 1.
        - Example: "1\n5\n1 2 3 4 5"
    - output_dict:
        - A Python dict that is a structured, formal description of the same test case.
        - If the problem statement contains multiple test cases, **do not** introduce T or any extra wrapper.
        - Example: {"n": 5, "list": [1, 2, 3, 4, 5]}
## Constraints
- All sizes (counts, lengths, number of operations, etc.) must be determined only from json_obj.
- All other values (elements of arrays, weights, edges, indices, etc.) must be generated randomly within a reasonable range and **strictly smaller than 10000**, while satisfying the problem's constraints at the same time.
- If the problem allows "no-solution" cases (e.g., the intended output is -1 when no solution exists), you should **bias your random generation towards test cases that admit at least one valid solution**, and explicitly construct values to satisfy any hidden feasibility conditions, so that the correct solution is not trivially always the "no-solution" output.
## Problem statement
{problem}
## Example json_obj
{example_json_obj}
\end{lstlisting}
\section{Environment Cases}
\label{app:environment_cases}

\subsection{Example Problem: 33\_C.Wonderful Randomized Sum}

\begin{longtable}{|>{\raggedright}m{3cm}|m{12cm}|}
\hline
\textbf{Name} & 33\_C. Wonderful Randomized Sum \\ \hline
\hline
\textbf{Problem Description} & Learn, learn and learn again — Valera has to do this every day. He is studying at mathematical school, where math is the main discipline. The mathematics teacher loves her discipline very much and tries to cultivate this love in children. That's why she always gives her students large and difficult homework. Despite that Valera is one of the best students, he failed to manage with the new homework. That's why he asks for your help. He has the following task. A sequence of n numbers is given. A prefix of a sequence is the part of the sequence (possibly empty), taken from the start of the sequence. A suffix of a sequence is the part of the sequence (possibly empty), taken from the end of the sequence. It is allowed to sequentially make two operations with the sequence. The first operation is to take some prefix of the sequence and multiply all numbers in this prefix by  - 1. The second operation is to take some suffix and multiply all numbers in it by  - 1. The chosen prefix and suffix may intersect. What is the maximum total sum of the sequence that can be obtained by applying the described operations? \\
\hline
\textbf{Generate Testcase} & 
\begin{lstlisting}[language=Python]
def generate_testcase(json_obj, output_format="str"):
    # Generate random test case for the problem
    n = int(json_obj.get("n", 10))
    numbers = random.sample(range(1, 100), n)
    
    if output_format == "dict":
        return {"n": n, "numbers": numbers}
    else:
        return f"{n}\n{' '.join(map(str, numbers))}"
\end{lstlisting} \\
\hline
\textbf{Output Requirement} & The first and the only line of the output should contain the answer to the problem. \\
\hline
\textbf{Difficulty Mapping} & 
\texttt{
\{ 
    "0": 0, "1": 1, "2": 2, "3": 2, "4": 3, "5": 4, 
    "6": 5, "7": 6, "8": 8, "9": 11, "10": 14, "11": 18, 
    "12": 23, "13": 30, "14": 39, "15": 51, "16": 67, 
    "17": 87, "18": 112, "19": 146, "20": 190, "21": 247, 
    "22": 318 
\}
} \\
\hline
\end{longtable}

\subsection{Example Problem: 1497\_D. Genius}

\begin{longtable}{|>{\raggedright}m{3cm}|m{12cm}|}
\hline
\textbf{Name} & 1497\_D. Genius \\
\hline
\textbf{Problem Description} &  Please note the non-standard memory limit.There are n problems numbered with integers from 1 to n. i-th problem has the complexity $c_i = 2^i$, tag $tag_i$ and score $s_i$.After solving the problem i it's allowed to solve problem j if and only if $IQ < |c_i - c_j|$ and $tag_i \neq tag_j$. After solving it your IQ changes and becomes $IQ = |c_i - c_j|$ and you gain $|s_i - s_j|$ points.Any problem can be the first. You can solve problems in any order and as many times as you want.Initially your IQ = 0. Find the maximum number of points that can be earned.\\
\hline
\textbf{Generate Testcase} & 
\begin{lstlisting}[language=Python]
def generate_testcase(json_obj, output_format="str"):
    # Generate a test case for the Genius problem
    n = int(json_obj.get("n", 10))
    tags = random.sample(range(1, n+1), n)
    scores = random.sample(range(1, 100), n)
    
    if output_format == "dict":
        return {"n": n, "tags": tags, "scores": scores}
    else:
        return f"{n}\n{' '.join(map(str, tags))}\n{' '.join(map(str, scores))}"
\end{lstlisting} \\
\hline
\textbf{Output Requirement} & For each test case print a single integer — the maximum number of points that can be earned. \\
\hline
\textbf{Difficulty Mapping} & 
\texttt{
\{ 
    "0": 0, "1": 2, "2": 3, "3": 4, "4": 7, "5": 10, 
    "6": 17, "7": 27, "8": 43, "9": 69, "10": 110, 
    "11": 176, "12": 281, "13": 450, "14": 721, "15": 1153
\}
} \\
\hline
\end{longtable}

\subsection{Example Problem: 1466\_B. Last Minute Enhancements}

\begin{longtable}{|>{\raggedright}m{3cm}|m{12cm}|}
\hline
\textbf{Name} & 1466\_B. Last Minute Enhancements \\
\hline
\textbf{Problem Description} & Athenaeus has just finished creating his latest musical composition and will present it tomorrow to the people of Athens. Unfortunately, the melody is rather dull and highly likely won't be met with a warm reception. His song consists of n notes, which we will treat as positive integers. The diversity of a song is the number of different notes it contains. As a patron of music, Euterpe watches over composers and guides them throughout the process of creating new melodies. She decided to help Athenaeus by changing his song to make it more diverse.Being a minor goddess, she cannot arbitrarily change the song. Instead, for each of the n notes in the song, she can either leave it as it is or increase it by 1.Given the song as a sequence of integers describing the notes, find out the maximal, achievable diversity. \\
\hline
\textbf{Generate Testcase} & 
\begin{lstlisting}[language=Python]
def generate_testcase(json_obj, output_format="str"):
    # Generate test case for the song diversity problem
    n = int(json_obj.get("n", 5))
    notes = random.sample(range(1, 10), n)
    
    if output_format == "dict":
        return {"n": n, "notes": notes}
    else:
        return f"{n}\n{' '.join(map(str, notes))}"
\end{lstlisting} \\
\hline
\textbf{Output Requirement} & For each test case, you should output a single line containing precisely one integer, the maximal diversity of the song, i.e. the maximal possible number of different elements in the final sequence. \\
\hline
\textbf{Difficulty Mapping} & 
\texttt{
\{ "0": 0,
      "1": 1,
      "2": 2,
      "3": 2,
      "4": 3,
      "5": 4,
      "6": 5,
      "7": 6,
      "8": 8,
      "9": 11,
      "10": 14,
      "11": 18,
      "12": 23,
      "13": 30,
      "14": 39,
      "15": 51,
      "16": 67,
      "17": 87,
      "18": 112,
      "19": 146,
      "20": 190,
      "21": 247,
      "22": 321,
      "23": 418
\}
} \\
\hline
\end{longtable}

\section{Training Detail}
\label{app:train_details}
\renewcommand{\arraystretch}{1.2} 
\begin{longtable}{|>{\raggedright}m{8cm}|m{5cm}|}
\hline
\textbf{Model Name} & \parbox[t]{8cm}{Qwen3-4B-Base \\ Qwen3-1.7B-Base} \\
\hline
\textbf{$K_{slope}$} & 10 \\
\hline
\textbf{$K_{zero}$} & 5 \\
\hline
\textbf{$K_{sat}$} & 5 \\
\hline
\textbf{Learning Rate} & 1e-6 \\
\hline
\textbf{Learning Rate Warmup Steps} & 20 \\
\hline
\textbf{Batch Size} & 64 \\
\hline
\textbf{Max Prompt Length} & 4096 \\
\hline
\textbf{Max Response Length} & 8192 \\
\hline
\textbf{Entropy Coefficient} & 0 \\
\hline
\textbf{Number of Environments per Step} & 64 \\
\hline
\textbf{Training Prompt Batch Size} & 64 \\
\hline
\textbf{Mini Batch Size} & 64 \\
\hline
\textbf{Reward Estimator} & grpo \\
\hline
\textbf{KL Loss Coefficient} & 0.0 \\
\hline
\textbf{Clip Ratio Low} & 0.2 \\
\hline
\textbf{Clip Ratio High} & 0.2 \\
\hline
\textbf{Temperature (Training)} & 1.0 \\
\hline
\textbf{Top P (Training)} & 1.0 \\
\hline
\textbf{Top K (Training)} & -1 \\
\hline
\textbf{Validation Temperature} & 0.6 \\
\hline
\textbf{Validation Top P} & 0.95 \\
\hline
\textbf{Number of Response per Prompt} & 8 \\
\hline
\end{longtable}

\section{Computational Cost and Efficiency}
\label{app:computational_cost}

\subsection{Synthesis Cost}
In our synthesis pipeline, we use approximately 70 million output tokens. This process involves two specific roles of agents:
\begin{itemize}
    \item The extraction of meta-information from the original programming problems.
    \item The synthesis of generating specific test cases under difficulty control.
\end{itemize}

\subsection{Training Overhead}
To investigate the training overhead of our method, we summarize the total training budget and resources for the Qwen3-4B-Base model in Table~\ref{tab:total_budget}. 

\begin{table*}[h]
\centering
\small
\setlength{\tabcolsep}{4pt}
\begin{tabular}{lcccccc}
\toprule
\textbf{Method} & \textbf{Total Steps} & \textbf{\shortstack{Prompts \\ / Step}} & \textbf{\shortstack{Responses \\ / Prompt}} & \textbf{\shortstack{Prompt Max \\ Tokens}} & \textbf{\shortstack{Response Max \\ Tokens}} & \textbf{\shortstack{Approx. GPU \\ Hours (H100)}} \\
\midrule
MATH     & 1300 & 64 & 8 & 4096 & 8192 & $59.33 \times 8$ \\
DeepMath & 1300 & 64 & 8 & 4096 & 8192 & $61.31 \times 8$ \\
\textbf{SCALER (Ours)}  & 1300 & 64 & 8 & 4096 & 8192 & $119.82 \times 8$ \\
\bottomrule
\end{tabular}
\caption{Summary of Total Training Budget and Resources for Qwen3-4B-Base.}
\label{tab:total_budget}
\end{table*}

As shown in Table~\ref{tab:total_budget}, SCALER consumes more GPU hours than the dataset-based baselines under identical training settings. However, the detailed breakdown of computational time per step in Table~\ref{tab:time_breakdown} demonstrates that this increase is primarily driven by the enhanced difficulty control strategy proposed in our method, rather than the overhead of environment interaction.

\begin{table*}[h]
\centering
\small
\renewcommand{\arraystretch}{1.2}
\begin{tabular}{l p{3.2cm} p{3.2cm} p{2.8cm} c}
\toprule
\textbf{Method} & \textbf{Environment Interaction \newline(Task/GT Gen) (s)} & \textbf{Model Training \newline(Rollout+Backward) (s)} & \textbf{Difficulty/Curation \newline Logic (s)} & \textbf{Total Time \newline per Step (s)} \\
\midrule
MATH     & \multicolumn{1}{c}{--} & 104.12 (100\%) & \multicolumn{1}{c}{--} & 104.12 \\
DeepMath & \multicolumn{1}{c}{--} & 108.56 (100\%) & \multicolumn{1}{c}{--} & 108.56 \\
\textbf{SCALER (Ours)}  & 37.17 (16.85\%) & 183.42 (83.15\%) & 0.0005 ($\approx$0\%) & 220.59 \\
\bottomrule
\end{tabular}
\caption{Breakdown of Computational Time per Training Step for Qwen3-4B-Base.}
\label{tab:time_breakdown}
\end{table*}

Based on the breakdown in Table~\ref{tab:time_breakdown}, we observe the following:
\begin{itemize}
    \item In the computational timeline of SCALER, environmental interaction accounts for only 16.85\% of the total runtime, while model training remains the dominant factor.
    \item Compared to dataset-based methods, the increase in model training time (183.42s vs. 104.12s) is significantly larger than the additional overhead introduced by environment interaction (37.17s). 
\end{itemize}

This increase in model training time is caused by longer response lengths and deeper reasoning steps, as detailed in Table~\ref{tab:response_length}. Under the paradigm of test-time scaling, more response tokens are naturally required to solve harder problems, which explicitly demonstrates the effectiveness of the difficulty control mechanism in our method.

\begin{table*}[h]
\centering
\small
\begin{tabular}{lccc}
\toprule
\textbf{Training Phase} & \textbf{MATH} & \textbf{DeepMath} & \textbf{SCALER} \\
\midrule
Early Stage(0--100 steps)    & 780.61   & 1,019.75 & 946.27 \\
Middle Stage(100--600 steps) & 1,375.46 & 1,266.22 & 3,164.77 \\
Later Stage(600--1300 steps) & 1,877.43 & 2,300.61 & \textbf{5,427.03} \\
\bottomrule
\end{tabular}
\caption{Comparison of Average Response Length between Dataset-based Methods and SCALER for Qwen3-4B-Base.}
\label{tab:response_length}
\end{table*}

\section{Stability and Statistical Significance Analysis}
\label{app:stability}

To demonstrate the stability and robustness of our proposed method, we conducted two additional independent training runs (with different random seeds) on the Qwen3-1.7B model. The results, summarized in Table~\ref{tab:stability}, show low variance across the independent runs, confirming that the performance improvements achieved by SCALER are statistically reliable rather than the result of random seed luck.

\begin{table*}[h]
\centering
\small
\setlength{\tabcolsep}{6pt}
\renewcommand{\arraystretch}{1.2}
\begin{tabular}{lcccccc}
\toprule
\textbf{Run} & \textbf{MATH-500} & \textbf{AMC23} & \textbf{AIME24} & \textbf{MMLU-Pro} & \textbf{BBEH} & \textbf{AVG} \\
\midrule
Run 1 (Reported) & 75.80 & 49.53 & 12.91 & 50.89 & 11.74 & 40.18 \\
Run 2            & 75.80 & 48.12 & 14.38 & 50.18 & 10.22 & 39.74 \\
Run 3            & 75.20 & 50.78 & 15.83 & 49.02 & 11.74 & 40.51 \\
\midrule
\textbf{Mean $\pm$ Std} & \textbf{75.60 $\pm$ 0.28} & \textbf{49.48 $\pm$ 1.09} & \textbf{14.37 $\pm$ 1.19} & \textbf{50.03 $\pm$ 0.77} & \textbf{11.23 $\pm$ 0.71} & \textbf{40.14 $\pm$ 0.32} \\
\bottomrule
\end{tabular}
\caption{Stability Analysis of SCALER on Qwen3-1.7B across 3 independent runs.}
\label{tab:stability}
\end{table*}

As shown in Table~\ref{tab:stability}, the overall average (AVG) across the three independent runs is $40.14 \pm 0.32$, which is highly consistent with the reported AVG of 40.18 in our main results. The standard deviation of the mean AVG is remarkably small relative to the overall score scale. Breaking down the performance by dataset, MATH-500, MMLU-Pro, and BBEH remain exceptionally stable. Although AMC23 and AIME24 exhibit slightly higher variances due to the inherently smaller size and higher difficulty of these evaluation sets, the variations remain well within an acceptable range. These results strongly support the reproducibility of SCALER's gains.

\clearpage

\bibliographystyle{plainnat}
\bibliography{main}

\end{document}